# PLANNING BY CASE-BASED REASONING BASED ON FUZZY LOGIC

Baghdad Atmani, Sofia Benbelkacem and Mohamed Benamina

Computer Science Laboratory of Oran (LIO)
Department of Computer Science, University of Oran, Algeria
BP 1524, El M'Naouer, Es Senia, 31 000 Oran, Algeria
{atmani.baghdad, sofia.benbelkacem, benamina.mohamed}@gmail.com

## ABSTRACT

*The treatment of complex systems often requires the manipulation of vague, imprecise and uncertain information. Indeed, the human being is competent in handling of such systems in a natural way. Instead of thinking in mathematical terms, humans describes the behavior of the system by language proposals. In order to represent this type of information, Zadeh proposed to model the mechanism of human thought by approximate reasoning based on linguistic variables. He introduced the theory of fuzzy sets in 1965, which provides an interface between language and digital worlds. In this paper, we propose a Boolean modeling of the fuzzy reasoning that we baptized Fuzzy-BML and uses the characteristics of induction graph classification. Fuzzy-BML is the process by which the retrieval phase of a CBR is modelled not in the conventional form of mathematical equations, but in the form of a database with membership functions of fuzzy rules.*

## 1. INTRODUCTION

The problem of planning and scheduling of tasks is one of the most complex problems in the field of Artificial Intelligence. The best-known situations include crisis management, production management, project management, robotics, medical, etc. The goal of planning is to provide a system (robotics, computer, human, ...) the capacity to reason to interact with its environment in an autonomous manner, in order to achieve the objectives that have been assigned. Planning is defined in terms of problems to be resolved and proposes a set of operators of change of State of the world, given a representation of the initial state of the world, and a goal to be attained, to give the means to a computer system to find a sequence of actions to be applied to the world to move from the initial state to a State that satisfies the goal to achieve.

Scheduling is organized in time a set of tasks. Historically, scheduling problems were discussed initially in the field of operational research (graph dynamic programming, linear programming, methods of combinatorial optimization theory), but quickly showed their limits in terms of expressiveness. Artificial intelligence and knowledge-based systems are then addressed the problem, renewing techniques through a richer representation of the domain knowledge (problems of satisfaction of constraints, constraints propagation algorithms, constraint programming languages). Among knowledge-based systems we looked on the reasoning from case (CBR).

The CBR based on artificial intelligence techniques is an approach to problem solving that uses past experiences to solve new problems by finding similar cases in its knowledge base and adapting them to the particular case. All the experiences form a case basis. Each case is represented by a knowledge experience. This experience is a lesson for the CBR system to solve problems of various kinds. The CBR consists of five phases : 1-Elaboration of the case. 2-Retrieval; 3-Adaptation; 4-Review and finally 5-Memory. For our project we are interested in the second phase: retrieval.

Therefore our contribution in this area is double, on the one hand offer a reactive planning module based on a CBR for the optimization of the scheduling, and on the other hand offer a classification induction graph [10] for the acceleration of the indexing of cases : remembering.

The classification issue is to assign the various observations to categories or predefined classes [16] [2]. In general classification methods consist in several stages. The most important step is to develop the rules of classification from a priori knowledge; It is the learning phase [11].

The classification by inductive learning finds its originality in the fact that humans often use it to resolve and to handle very complex situations in their lives daily [19]. However, the induction in humans is often approximate rather than exact. Indeed, the human brain is able to handle imprecise, vague, uncertain and incomplete information [18]. Also, the human brain is able to learn and to operate in a context where uncertainty management is indispensable. In this paper, we propose a Boolean model of fuzzy reasoning for indexing the sub-plans, based on characteristics of the classification by inductive learning in humans [22].

This article is structured as follows. Section 2 presents a state of the art of work about planning and data mining. Section 3 is devoted to the construction of the base of the case. In section 4 we discuss classification by inductive learning from data and in particular the induction of rules by graph of induction. In section 5 we introduce Boolean modeling [1]. Fuzzy logic is discussed in section 6. Section 7 presents results of experimentation. Finally, we present the guidance of our contribution and experimentation and we conclude in section 8.

## 2. STATE OF THE ART

We present previous work which have combined planning and data mining.

Kaufman and Michalski [15] propose an approach that involves the integration of various processes of learning and inference in a system that automatically search for different data mining tasks according to a high-level plan developed by a user. This plan is specified in a language of knowledge production, called KGL (Knowledge Generation Language).

Kalousis and al. [14] propose a system that combines planning and metalearning to provide support to users of a virtual laboratory data mining. The addition of meta-learning to planning based data mining support will make the planner adaptive to changes in the data and capable of improving its advice over time. Planner based on knowledge is based on ontology of data mining workflow for planning knowledge discovery and determine the set of valid operator for each stage of the workflow.

Záková and al. [20] have proposed a methodology that defines a formal conceptualization of the types of knowledge and data mining algorithms as well as a planning algorithm that extracts the constraints of this conceptualization according to the requirements given by the user. The task of building automated workflow includes the following steps: converting the task of knowledge discovery into a planning problem, plan generation using a planning algorithm, storing the generated abstract workflow in form of semantic annotation, instantiating the abstract workflow with specific configurations of the required algorithms and storing the generated workflow.

Fernández and al. [12] presented a tool based on automated planning that helps users, not necessarily experts on data minig, to perform data mining tasks. The starting point will be a definition of the data mining task to be carried out and the output will be a set of plans. These plans are executed with the data mining tool WEKA [19] to obtain a set of models and statistics. First, the data mining tasks are described in PMML (Predictive Model Markup Language). Then, from the PMML file a description of the planning problem is generated in PDDL (the standard language in the planning community). Finally, the plan is being implemented in WEKA (Waikato Environment for Knowledge Analysis).

## 3. CONSTRUCTION OF THE CASE BASE

Case-based reasoning is one of the currently most widely used artificial intelligence techniques. Reasoning from cases is to solve a new problem, called problem target, using a set of problems already solved. A source case refers to an episode from problem solving and a case one basis together cases sources [3]. A case consists of two parts: the problem and the solution part. The problem part is described by a set of indices that determine in what situation a case is applicable. Case-based reasoning process generally operates under five sequential phases: development, remembering (or indexing), adaptation, the revision and learning.

Scheduling based on the case is a planning approach that is based on a particular aspect of human behaviour. Generally, the man does not generate plans (calendar) entirely new from basic operations; he uses his experiences (success or failure) to help solve new problems that arise to him. Establish a schedule returned to try to synthesize a solution plan, by reusing the best plans already produced in similar situations and changing to adapt to the new situation. Planning from case, a scheduling problem is the specification of an initial state and a goal to achieve. A solution is a plan for achieving the purpose starting from the initial state.

For a system of case-based reasoning to work, it must start from a certain number of cases constituting the basis of cases. These cases should cover the target area the best possible so interesting solutions are found. Take the example of the treatment of a disease reportable: tuberculosis. The treatment of tuberculosis differs depending on the patient's age and various other factors. We are building the basis of cases passing through four steps: the project description, modeling of the project by a graph or, the generation of the plans, the construction and representation of cases [6].

The description of the project is to represent the sequence of tasks or actions as an array.

Then, a graph or is generated from the project. The graph is a graph whose nodes represent tasks or arcs represent the relationships between tasks. The relationship between the tasks being to satisfy constraints [4]. Constraints are criteria that may be taken into account in the development of sub-plans. The choice of a solution plan depends on several criteria: time, probability, and the cost. However a task represents the action carried out for a period of time [17].

After the construction of the graph, we and/or apply planning algorithms to determine the possible plans. The scheduling algorithm we use is based on a course of the graph or back chaining. It is to find possible paths between the initial node and the final node of the graph and/or. The algorithm stops when the sought initial node is found [8].

To build the case, we will link a duration, a probability and a cost to each plan obtained in the previous step according to its tasks. Therefore cases will be represented by descriptors (duration, cost, likelihood) that describe the problem part and the corresponding plane that represents the solution part. Table 1 presents the basis of 14 cases.

# 4. CLASSIFICATION OF PLANS BY INDUCTIVE LEARNING

Establish a plan, means being able to associate the subplan to a number of indices presented by situations. In this type of problem, it identifies three essentials: problems, plans and indices. Problems are the population, indices are the descriptions of the problems and plans are the classes. It is assumed that there is a correct classification, meaning that there is an application that associates any scheduling problem with a plan. Learn how to develop a plan, is to associate a plan already drawn up a list of indices. To formalize this connection, we will use the following notations: $\Omega = \{w_1, w_2,...,w_n\}$ to refer to a population of *n* scheduling problems. $G=\{g_1,g_2,...,g_d\}$ for all *d* descriptions (indices of the problem) and $Q=\{q_1,q_2,...,q_m\}$ for all the plans *m*.

Is $\Omega$ a population of individuals affected by the problem of classification. This population is a special attribute called noted class attribute *is* associated. The variable *Y* is called the area of variable statistics endogenous or simply class. At each individual w may be associated with its class *Y*(w). They say that the function *is* takes its values in the set of labels Q, called also whole classes. For example, if the $\Omega$ population is diabetic patients and *is* the result of the identification of diabetes type 1 noted $q_1$, and type 2 noted $q_2$; then *Y*(w) will be the result of the identification of the type of diabetes the patient w [1].

The determination of the classification model $\varphi$ is related to the assumption that the values taken by the variable *Y* are not random, but certain specific situations that can characterize [22]. For this the expert in the field concerned establishes a priori list of *p* variable statistical called variables exogenous and rated $X= \{X_1, X_2,..., X_p\}$. These variables are also called predictive attributes or explanatory. The value taken by a variable exogenous $X_j$ is called modality or value of attribute $X_j$ of the problem w. We mean by $l_j$ the number of terms that a variable $X_j$ can receive. To illustrate this notation, consider the problem of planning. A problem can be described, for example, by three exogenous variables:

$X_1$: Duration, which can take three values  $x_1^1$=*Courte*, $x_1^2$=*Normale*, $x_1^3$=*Longue*

$X_2$: Probability, which can take three values $x_2^1$=*Incertain*, $x_2^2$=*Douteux*, $x_2^3$=*Certain*

$X_3$: Cost, which can take three values $x_3^1$=*Faible*, $x_3^2$=*Raisonnable*, $x_3^3$=*Elevé*

Inductive learning aims to seek a classification model $\varphi$ allowing for a new case *w*, for which we do not know the class *Y (w)* but we know the State of all of its variables exogenous to predict this value throughj. The development of $\varphi$ requires in the population $\Omega$ two samples graded $\Omega_A$ and $\Omega_T$. The first said of learning will be used for construction and the second said to test will be used to test the validity of $\varphi$. Thus, for any case *w*, we assume known both its values *X (w)* in the space of representation and its class *Y (w)* space labels Q.

Population $\Omega_A$ cases, taken into account for classification is nothing more than a sequence of *n* case *w* $_i$ (situations) with their plan corresponding *Y*($w_i$). Suppose that the $\Omega_A$ sample is composed of 14 situations (table 1):

Table 1. Example of a learning sample.

| Ω | Y(ω) | $X_1(ω)$ | $X_2(ω)$ | $X_3(ω)$ | Ω | Y(ω) | $X_1(ω)$ | $X_2(ω)$ | $X_3(ω)$ |
|---|---|---|---|---|---|---|---|---|---|
| $ω_1$ | Plan1 | 75 | 0,70 | 70 | $ω_8$ | Plan1 | 64 | 0,40 | 65 |
| $ω_2$ | Plan2 | 80 | 0,80 | 90 | $ω_9$ | Plan1 | 65 | 0,80 | 75 |
| $ω_3$ | Plan2 | 85 | 0,85 | 85 | $ω_{10}$ | Plan2 | 51 | 0,10 | 80 |
| $ω_4$ | Plan2 | 72 | 0,20 | 95 | $ω_{11}$ | Plan2 | 55 | 0,50 | 70 |
| $ω_5$ | Plan1 | 79 | 0,69 | 70 | $ω_{12}$ | Plan1 | 49 | 0,52 | 80 |
| $ω_6$ | Plan1 | 71 | 0,70 | 90 | $ω_{13}$ | Plan1 | 58 | 0,81 | 80 |
| $ω_7$ | Plan1 | 63 | 0,30 | 78 | $ω_{14}$ | Plan1 | 40 | 0,90 | 96 |

Supervised inductive learning intends to provide tools for extracting the classification model based on the information available on the sample of learning φ. The general process of inductive learning includes typically three steps that we summarize below:

1. **Development of the model:** This is the step that uses a sample of noted learning $Ω_A$, which all individuals $w_i$ are described in a space of representation and belong to one of the *m* classes denoted $c_j$, *j*= 1,...,*m*. It is building the application φ which allows calculating the class from representation.

2. **Validation of the model:** This is to verify on a sample test $Ω_T$ and which we know for each of its individuals, representation and the class, if the classification model φ from step previous gives of the class expected.

3. **Generalization of the model:** This is the stage which is to extend the application of the model to all individuals of the population Ω.

## 5. BOOLEAN MODELING OF THE INDUCTION GRAPH

In this section, we present the principles of construction, by Boolean modelling [1,5,9,23,24], of induction graphs in the problems of discrimination and classification [1,2] : we want to explain the class taken by one variable to predict categorical *Y*, attribute class or endogenous variable; from a series of variables $X_1, X_2,..., X_p$, say variable predictive (descriptors) or exogenous, discrete or continuous. According to the terminology of machine learning, we are therefore in the context of supervised learning. The general process of learning than the cellular system *CASI* (Cellular Automata for Symbolic Induction) [1] applies to a population is organized on three stages:

1) Boolean modeling of the induction graph;

2) Generation of the rules for cases indexing;

3) Validation and generalization;

Figure 1 summarizes the general diagram of the Boolean modeling process in the CASI system.

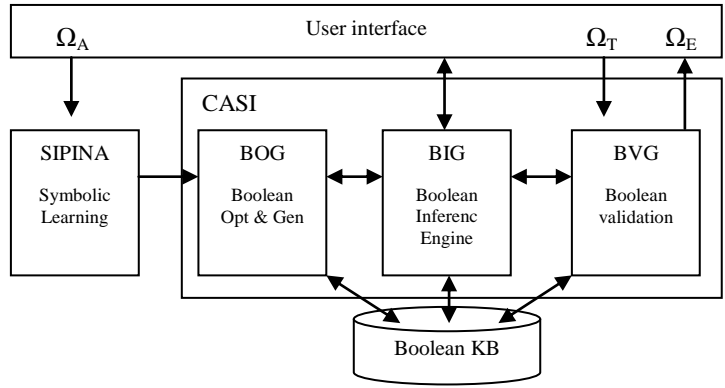

Figure 1. General diagram of the system CASI

From the sample $\Omega_A$ we begin the symbolic treatment for the construction of the induction graph (method *SIPINA* [21] [22].

1) Choose the extent of uncertainty (Shannon or quadratic);

2) Initialize the parameters $\lambda$, $\mu$ and the initial partition $S_0$;

3) Use the *SIPINA* method to pass partition $S_t$ to $S_{t+1}$ and generate the graph of induction.

4) Finally, generation of prediction rules.

Method *SIPINA* [21] algorithm is a non tree heuristic for the construction of a graph of induction. Its principle is to generate a succession of scores by merger or breakup of the nodes of the graph. In what follows we describe the process on the fictional example of table 1. Suppose our sample of learning $\Omega_A$ consists of 14 cases of scheduling which are divided into two classes *plan1* - *plan2* (see table 1). The initial partition $S_0$ has one $s_0$ noted element, which includes the entire sample learning with 9 situations belonging to the class *plan1* and 5, class *plan2*. The next partition $S_1$ is generated by the variable $X_1$ after discretization and individuals in each node $s_i$ are defined as follows: $s_1=\{\omega\in\Omega_A/X_1(\omega)=Longue$ pour $X_1(\omega)>=72\}$, $s_2=\{\omega\in\Omega_A/X_1(\omega)=Normale$ pour $X_1(\omega)>=60$ et $X_1(\omega)<72\}$ and $s_3=\{\omega\in\Omega_A/X_1(\omega)=Courte$ pour $X_1(\omega)<60\}$.

As well as in the $s_0$ node, there are in $s_1$, $s_2$ and $s_3$, individuals of the *plan1* and *plan2* classes. The figure 2 summarizes the steps of construction of $s_0$, $s_1$, $s_2$ and $s_3$. The $S_1$ partition, the process is repeated looking for a $S_2$ score which would be better.

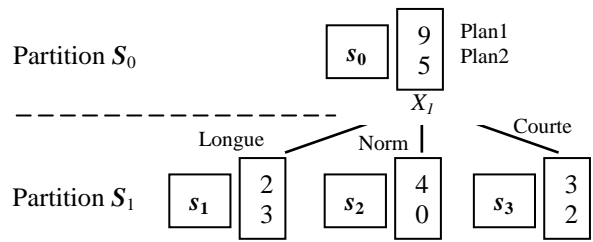

Figure 2. Construction of the nodes s0, s1, s2 and s3

To illustrate the architecture and the operating principle of the *BIG* module, we consider figure 2 with the $S_0 = (s_0)$ partitions and $S_1 = (s_1, s_2, s_3)$. Figure 3 shows how the knowledge extracted from this graph database is represented by the *CELFACT* and *CELRULE* layers. Initially, all entries in cells in the *CELFACT* layer are passive ($E = 0$), except for those who represent the initial basis of facts ($E = 1$).

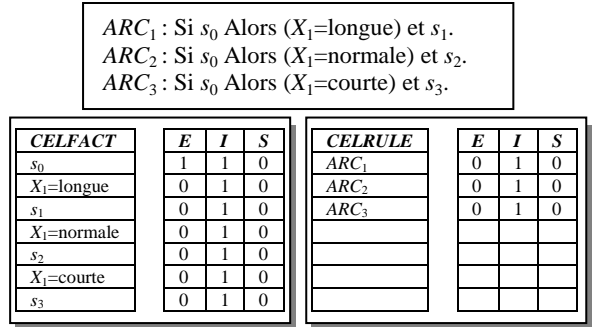

$ARC_1$ : Si $s_0$ Alors ($X_1$=longue) et $s_1$.
$ARC_2$ : Si $s_0$ Alors ($X_1$=normale) et $s_2$.
$ARC_3$ : Si $s_0$ Alors ($X_1$=courte) et $s_3$.

| CELFACT | E | I | S |
|---|---|---|---|
| $s_0$ | 1 | 1 | 0 |
| $X_1$=longue | 0 | 1 | 0 |
| $s_1$ | 0 | 1 | 0 |
| $X_1$=normale | 0 | 1 | 0 |
| $s_2$ | 0 | 1 | 0 |
| $X_1$=courte | 0 | 1 | 0 |
| $s_3$ | 0 | 1 | 0 |

| CELRULE | E | I | S |
|---|---|---|---|
| $ARC_1$ | 0 | 1 | 0 |
| $ARC_2$ | 0 | 1 | 0 |
| $ARC_3$ | 0 | 1 | 0 |

Figure 3. Boolean partitions modeling $S_0$ and $S_1$

In figure 4 are, respectively, represented the impact of input matrices $R_E$ and exit $R_S$ the Boolean model.

· the relationship entry, denoted i $R_E$ j, is formulated as follows: $\forall i \in \{1,..., l\}, \forall j \in \{1,..., r\}$, if (the fact i ∈ to the premise of the j rule) then $R_E(i, j) \leftarrow 1$.

· the relationship of output, denoted i $R_S$ j, is formulated as follows: $\forall i \in \{1,..l\}, \forall j \in \{1,..., r\}$, if (the fact i ∈ the conclusion of rule j) then $R_S(i, j) \leftarrow 1$.

| $R_E$ | $ARC_1$ | $ARC_2$ | $ARC_3$ |
|---|---|---|---|
| $s_0$ | 1 | 1 | 1 |
| $X_1$=longue | 0 | 0 | 0 |
| $s_1$ | 0 | 0 | 0 |
| $X_1$=normale | 0 | 0 | 0 |
| $s_2$ | 0 | 0 | 0 |
| $X_1$=courte | 0 | 0 | 0 |
| $s_3$ | 0 | 0 | 0 |

| $R_S$ | $ARC_1$ | $ARC_2$ | $ARC_3$ |
|---|---|---|---|
| $s_0$ | 0 | 0 | 0 |
| $X_1$=longue | 1 | 0 | 0 |
| $s_1$ | 1 | 0 | 0 |
| $X_1$=normale | 0 | 1 | 0 |
| $s_2$ | 0 | 1 | 0 |
| $X_1$=courte | 0 | 0 | 1 |
| $s_3$ | 0 | 0 | 1 |

Figure 4. Input/output incidences matrices

Incidence matrices $R_E$ and $R_S$ represent the relationship *input/output* of the facts and are used in *forward-chaining* [1] [9]. You can also use $R_S$ as relationship of input and $R_E$ as relationship of output to run a rear chaining inference. Note that no cells in the vicinity of a cell that belongs to *CELFACT* (at *CELRULE*) does not belong to the layer *CELFACT* (at *CELRULE*).

The dynamics of the cellular automaton *BIG* [1,23], to simulate the operation of an *Inference engine* uses two functions of transitions $\delta_{fact}$ and $\delta_{rule}$, where $\delta_{fact}$ corresponds to the phase of *assessment*, *selection* and *filtering*, and $\delta_{rule}$ corresponds to the *execution* phase [1,24]. To set the two functions of transition we will adopt the following notation: EF, IF and SF to designate CELFACT_E, _I and _S; Respectively ER, IR and SR to designate *CELRULE*_E, _I and _S.

-The transition function $\delta_{fact}$: (EF, IF, SF, ER, IR, SR) ➔ (EF, IF, **EF**, **ER+($R_E^T$·EF)**, IR, SR)

-The transition function $\delta_{rule}$ : (EF, IF, SF, ER, IR, SR) ➔ (**EF+($R_S$·ER)**, IF, SF, ER, IR, **§ER**)

Where $R_E^T$ matrix is the transpose of $R_E$ and where §ER is the logical negation of ER. Operators + and · used are respectively the or and the and logical.

We consider G0 initial configuration of our cellular automaton (see figure 4), and Δ = δrule o δfact the global transition function: Δ (G0) = G1 if δfact (G0) = G'0 and δrule (G'0) = G1. Suppose that G = {G0, G1,..., Gq} is the set of Boolean PLC configurations. Discrete developments plc, from one generation to another, is defined by the sequence G0, G1,..., Gq, where Gi+1=Δ(Gi) [1,23,24].

## 6. FUZZY BOOLEAN MODELING

According to Lotfi Zadeh [19], founder of fuzzy logic, the limits of the classical theories applied in artificial intelligence come because they require and manipulate only accurate information. Fuzzy logic provides approximate reasoning modes rather than accurate. It is mainly the mode of reasoning used in most cases in humans.

According to Zadeh, fuzzy logic is fuzzy sets theory which is a mathematical theory, whose main objective is the modeling of the vague and uncertain of the natural language concepts. Thus, it avoids the inadequacies of the traditional theory regarding the treatment of this kind of knowledge. The fundamental characteristic of a classic set is rigid boundary between two classes of elements: those who belong to all and those who do not belong to this set. they belong rather to its complement. The relationship of belonging is represented in this case by a μ function that takes truth values {0,1} pair. Thus, the membership of a classic set A function is defined by:

$$\mu_A(x) = \begin{cases} 1 \; if \; x \in A \\ 0 \; if \; x \notin A \end{cases}$$

This means that an element $x$ in A ($\mu_A(x)=1$) or not ($\mu_A(x)=0$). However, in many situations, it is sometimes ambiguous whether $x$ belongs or not to A.

As an example for the definition of the membership functions, it takes the variable $X_3$ = Coût of table 2. According to the (logical Boolean) classical logic, which allows for variables that two values 0 and 1, all costs less than 40 are considered low, and those over 70 as high. However such logic of classification does not make sense. Why a cost of 75 is considered higher? in reality such a passage is done gradually. Fuzzy logic variables may take any values between 0 and 1 to take account of this reality.

In the simplest case, one can distinguish three values *Faible*, *Raisonnable*, and *Elevé*, of the language variable "Coût" which forms three fuzzy sets (figure 5). Thus, a cost of 35 belongs with a factor of belonging μ = 0.75 across ' Faible' and with μ = 0.25 to all 'Raisonnable'. Obviously, the choice characterizing the trapezoidal shape of the membership function is quite arbitrary and must take account of the particular circumstances. Often, it is necessary to introduce a subdivision more fine, e.g. 5 values «très faible», «faible», «raisonnable», «élevé» and «très élevé» for the language variable 'Cost', thus forming 5 sets. Thus a cost of 35 belongs, with μ = 0.25, across ' faible '.

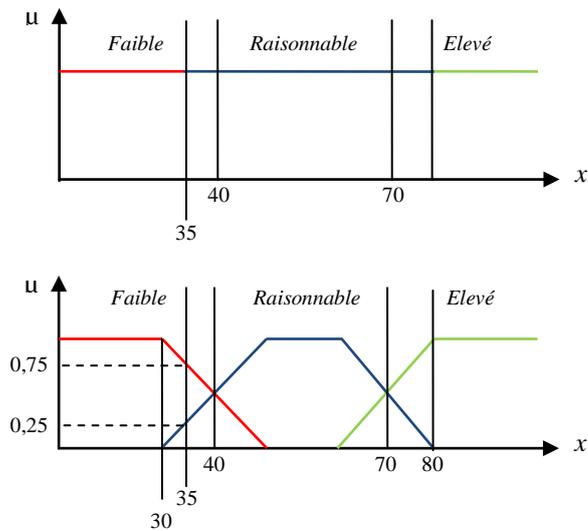

Figure 5. Classification of the cost according to the logic fuzzy

Suppose now that from the induction graph obtained with the method SIPINA we generated five rules $R_1$, $R_2$, $R_3$, $R_4$ and $R_5$ of classification that we'll use for the indexing of cases. The following illustration shows the Boolean modelling of extracted knowledge base.

$R_1$ : Si ($X_1$=Longue) et ($X_3$=Faible) Alors *plan1*.
$R_2$ : Si ($X_1$=Longue) et ($X_3$=Elevé) Alors *plan2*.
$R_3$ : Si ($X_1$=Normale) Alors *plan1*.
$R_4$ : Si ($X_1$=Courte) et ($X_2$=*Incertain*) Alors *plan2*.
$R_5$ : Si ($X_1$=Courte) et ($X_2$=*Douteux*) Alors *plan1*.

| CELFACT | E | I | S |
|---|---|---|---|
| $X_1$=Longue | 1 | 1 | 0 |
| $X_1$=Normale | 0 | 1 | 0 |
| $X_1$=Courte | 0 | 1 | 0 |
| $X_2$=Incertain | 0 | 1 | 0 |
| $X_2$=Douteux | 0 | 1 | 0 |
| $X_2$=Certain | 0 | 1 | 0 |
| $X_3$=Faible | 0 | 1 | 0 |
| $X_3$=Raisonnable | 0 | 1 | 0 |
| $X_3$=Elevé | 0 | 1 | 0 |
| Plan1 | 0 | 1 | 0 |
| Plan2 | 0 | 1 | 0 |

| CELRULE | E | I | S |
|---|---|---|---|
| $R_1$ | 0 | 1 | 0 |
| $R_2$ | 0 | 1 | 0 |
| $R_3$ | 0 | 1 | 0 |
| $R_4$ | 0 | 1 | 0 |
| $R_5$ | 0 | 1 | 0 |

| $R_E$ | $R_1$ | $R_2$ | $R_3$ | $R_4$ | $R_5$ |
|---|---|---|---|---|---|
| $X_1$=Longue | 1 | 1 | 0 | 0 | 0 |
| $X_1$=Normale | 0 | 0 | 1 | 0 | 0 |
| $X_1$=Courte | 0 | 0 | 0 | 1 | 1 |
| $X_2$=Incertain | 0 | 0 | 0 | 1 | 0 |
| $X_2$=Douteux | 0 | 0 | 0 | 0 | 1 |
| $X_2$=Certain | 0 | 0 | 0 | 0 | 0 |
| $X_3$=Faible | 1 | 0 | 0 | 0 | 0 |
| $X_3$=Raisonnable | 0 | 0 | 0 | 0 | 0 |
| $X_3$=Elevé | 0 | 1 | 0 | 0 | 0 |
| Plan1 | 0 | 0 | 0 | 0 | 0 |
| Plan2 | 0 | 0 | 0 | 0 | 0 |

| $R_E$ | $R_1$ | $R_2$ | $R_3$ | $R_4$ | $R_5$ |
|---|---|---|---|---|---|
| $X_1$=Longue | 0 | 0 | 0 | 0 | 0 |
| $X_1$=Normale | 0 | 0 | 0 | 0 | 0 |
| $X_1$=Courte | 0 | 0 | 0 | 0 | 0 |
| $X_2$=Incertain | 0 | 0 | 0 | 0 | 0 |
| $X_2$=Douteux | 0 | 0 | 0 | 0 | 0 |
| $X_2$=Certain | 0 | 0 | 0 | 0 | 0 |
| $X_3$=Faible | 0 | 0 | 0 | 0 | 0 |
| $X_3$=Raisonnable | 0 | 0 | 0 | 0 | 0 |
| $X_3$=Elevé | 0 | 0 | 0 | 0 | 0 |
| Plan1 | 1 | 0 | 1 | 0 | 1 |
| Plan2 | 0 | 1 | 0 | 1 | 0 |

Figure 6. Boolean modelling of extracted knowledge base

## 6.1. Boolean Fuzzification of Exogenous Variables

Fuzzy-BML modelling deals with the fuzzy input variables and provides results on output variables themselves blurred. Fuzzification, illustrated by the following example, is the step that consists of fuzzy quantification of actual values of a language variable.

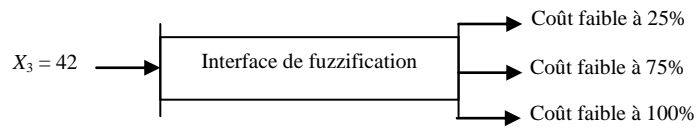

Fuzzifier to: the universe of discourse, i.e. a range of possible variations of the corresponding entry. A partition interval fuzzy from this universe, for the identification of the cost we partitioned space of X3 to 7 with a Boolean modeling on 3 bits of 000 to 110. Finally, the duties of membership classes.

## 6.2. Boolean Defuzzification

Output the Fuzzy-BML modeling cannot communicate to the user of the fuzzy values. The role of the defuzzification is therefore to provide accurate values. During this step, the system will perform tests to define the range of proven goal. This test will depend on the number of rules candidates and the de facto number of each rule that participated in the inference according to the following principle:

• Cases for a single rule and a single fact: "if then conclusion.

CELFACT _I (conclusion) = minimum (CELFACT_I (fact), CELRULE(rule) _I).

• Cases for a single rule with several facts: ' If fait1 and fait2 and... .' then conclusion» : CELFACT_I (conclusion) = minimum (CELFACT_I (fait1), CELFACT(fait2) _I, ...).

The 'minimum' operator in Boolean logic represents the "and logical."

• Several rules:

CELFACT _I (goal) = maximum (CELRULE_I (rule1), CELRULE_I (rule2),...).

The 'maximum' operator in Boolean logic represents the "logical or". Figure 7 shows the Boolean principle adopted by the Fuzzy-BML modeling.

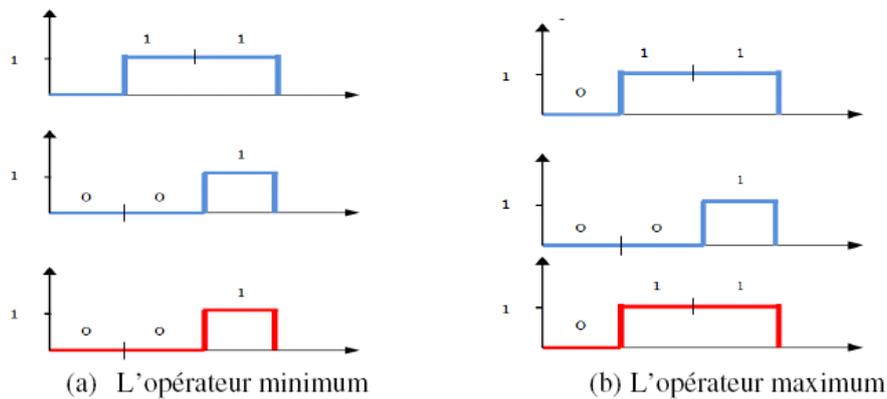

Figure 7. Boolean for the defuzzification operator

## 7. EXPERIMENTATION

We have evaluated our approach on a case basis about the treatment of tuberculosis. The case basis contains actual cases collected from the CHU of Oran [7]. The problem part of cases is described by three descriptors given in Table 2 and the solution part is given in the form of a treatment plan Y which takes its values in the set of plans C={T1, T2, T3, T4}.

Table 2. Values of descriptors.

| Descriptors | Meaning | Values |
|---|---|---|
| $X_1$ | Age | < 20, 20-30, 30-40, 40-50, >50 |
| $X_2$ | Weight | 30-39, 39-54, 54-70, >70 |
| $X_3$ | Antecedent | NT, T |

To compare the proposed approach with other methods, we have applied the k-NN [13], the decision tree and the Fuzzy-BML on the same case base. We show in Table 3 the rate of correctly classified instances with each method using the supervised mode of discretization.

Table 3. Results of experimentation.

| k-NN | Decision tree | Fuzzy-BML |
|---|---|---|
| 66 % | 73 % | 81 % |

The rate of correctly classified instances is 66 % with k-NN, 73 % with decision tree and 81 % with Fuzzy-BML. From the obtained results, we note that the Fuzzy-BML method has provided better results with a rate of 81 % of well classified instances.

## 8. CONCLUSIONS AND PERSPECTIVES

Several competing motivations have led us to define a Boolean model for CBR knowledge base systems. Indeed, we have not only desired experiment with a new approach to indexing of cases by decision tree, but we also wanted improve modeling of the vague and uncertain of the natural language concepts. When it comes to planning guided by CBR, we must go through the following steps:

- Build the base of cases by planning tools;

- Construct the graph of induction by symbolic learning and extract the rules;

- Import knowledge base in the platform WCSS [5].

- Launch the Boolean fuzzification [9] ;

- Launch the inference blurred for indexing in the basis of the cases;

- Finally, and if necessary run the Boolean defuzzification.

For the calculation of the similarity in the retrieval (cases indexing) phase, typically used k-nearest neighbours. So we compared our Fuzzy Boolean Model with k-nearest neighbours (k-NN) and decision tree. We noticed that the indexing of cases for the choice of a plan is significantly better with Fuzzy-BML. Finally, we can say that the structure of the cases that we have used is quite simple. We have described the part problem of cases by age, weight and a antecedent. By adding other constraints could subsequently used a slightly more complex representation. As a future perspective of this work, we propose to improve the other steps of the CBR process for the proposed approach

## REFERENCES

[1] Atmani, B. & Beldjilali, B. (2007) "Knowledge Discovery in Database: Induction Graph and Cellular Automaton", *Computing and Informatics Journal*, Vol. 26, No. 2, pp171-197.

[2] Atmani, B. & Beldjilali, B. (2007) "Neuro-IG: A Hybrid System for Selection and Eliminatio of Predictor Variables and non Relevant Individuals", *Informatica, Journal International*, Vol. 18, No. 2, pp163-186.

[3] Badra, F. (2009) « Extraction de connaissances d'adaptation en raisonnement à partir de cas », Thèse de doctorat, Université Henri Poincaré Nancy 1.

[4] Baki, B. (2006) « Planification et ordonnancement probabilistes sous contraintes temporelles », Thèse de doctorat, Université de CAEN.

[5] Benamina, B. & Atmani, B. (2008) « WCSS: un système cellulaire d'extraction et de gestion des connaissances », *Troisième atelier sur les systèmes décisionnels*, 10 et 11 octobre 2008, Mohammadia – Maroc, pp223-234.

[6] Benbelkacem, S., Atmani, B. & Mansoul, A. (2012) « Planification guidée par raisonnement à base de cas et datamining : Remémoration des cas par arbre de décision », *aIde à la Décision à tous les Etages* (Aide@EGC2012), pp62–72.

[7] Benbelkacem, S., Atmani & B., Benamina, M. (2013) "Treatment tuberculosis retrieval using decision tree", *The 2013 International Conference on Control, Decision and Information Technologies* (CoDIT'13).

[8] Benbelkacem, S., Atmani & B., Benamina, M. (2013) « Planification basée sur la classification par arbre de décision », *Conférence Maghrébine sur les Avancées des Systèmes Décisionnels* (ASD'2013).

[9] Beldjilali, A. & Atmani, B. (2009) « Identification du type de diabète par une approche cellulo-floue », *Quatrième atelier sur les systèmes décisionnels*, 10 et 11 novembre 2009, Jijel –Algérie, pp203-218.

[10] Breiman, L., Friedman, J. H., Olshen, R. A. & Stone, C. J. (1984) " Classification and regression and trees", Technical report, Wadsworth International, Monterey, CA.

[11] Fayyad, U., Shapiro, G.P. & Smyth, P. (1996) "The KDD process for extraction useful knowledge from volumes data", *Communication of the ACM*.

[12] Fernandez, S., la Rosa, T. D. , Fernandez, F., Suarez, R., Ortiz, J., Borrajo, D. & Manzano, D. (2009) "Using automated planning for improving data mining processes", *The Knowledge Engineering Review*.

[13] Guo, G., Wang, H., Bell, D., Bi, Y. & Greer, K. (2003) "Knn model-based approach in classification", *International Conference on Ontologies, Databases and Applications of Semantics* (ODBASE 2003), pp986– 996.

[14] Kalousis, A., Bernstein, A. & Hilario, M. (2008) "Meta-learning with kernels and similarity functions for planning of data mining workflows. ICML/COLT/UAI 2008, *Planning to Learn Workshop* (PlanLearn), pp23–28.

[15] Kaufman, K. & Michalski, R. (1998) "Discovery planning: Multistrategy learning in data mining", *Fourth International Workshop on Multistrategy Learning*, pp14–20.

[16] Kodratoff, Y. (1997) "The extraction of knowledge from data, a new topic for the scientific research", Magazine electronic READ.

[17] Régnier, P. (2005) « Algorithmes pour la planification. Habilitation à diriger des recherches », Université Paul Sabatier.

[18] Zadeh, Lotfi A (1968) "Probability measures of fuzzy events. Journal of Mathematical Analysis and Applications", Vol. 23, pp421–427.

[19] Zadeh, Lotfi A (1997) «Some Reflections on the Relationship Between AI and Fuzzy Logic: A Heretical View», IJCAI, Springer, pp1-8.

[20] Zakova, M., Kremen, P., Zelezny, F. & Lavrac, N. (2008) "Using ontological reasoning and planning for data mining workflow composition", Proc. of ECML/PKDD *workshop on Third Generation Data Mining: Towards Service-oriented*, pp35–41.

[21] Zighed, D.A.: SIPINA for Windows, ver 2.5. Laboratory ERIC, University of Lyon, 1996.

[22] Zighed, D.A. & Rakotomalala, R. (2000) "Graphs of induction, Training and Data Mining", Hermes Science Publication, pp 21-23.

[23] Brahami, M., Atmani, B. & Matta, N. (2013) "Dynamic knowledge mapping guided by data mining: Application on Healthcare", *Journal of Information Processing Systems*, Vol. *9* , No 1, pp 01-30.

[24] Barigou, F., Atmani, B. & Beldjilali, B. (2012) "Using a cellular automaton to extract medical information from clinical reports", *Journal of Information Processing Systems*, Vol. 8, No 1, pp67-84.